%% file: template.tex
\documentclass{article}
\usepackage{arxiv}
\usepackage{hyperref}       % hyperlinks
\usepackage{booktabs, array}       % professional-quality tables
\usepackage{amsfonts, amsmath, amssymb}       % blackboard math symbols
\usepackage{graphicx, multirow}
\usepackage{float}
\usepackage[table]{xcolor}
\title{The Dice loss in the context of missing or empty labels: introducing $\Phi$ and $\epsilon$}
\renewcommand{\undertitle}{This is an accepted MICCAI 2022 preprint}
\renewcommand{\headeright}{MICCAI 2022}
\renewcommand{\shorttitle}{The Dice loss in the context of missing or empty labels}
\author{
    \href{https://orcid.org/0000-0002-6123-7760}{\includegraphics[scale=0.06]{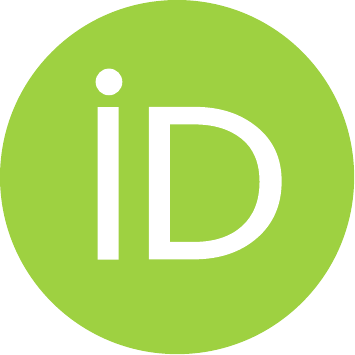}\hspace{1mm}Sofie Tilborghs}\thanks{S. Tilborghs and J. Bertels have contributed equally to this work.}\\
	Processing Speech and Images\\
	Department of Electrical Engineering\\
	KU Leuven, Belgium\\
	\texttt{sofie.tilborghs@kuleuven.be}\\
	\And
    \href{https://orcid.org/0000-0001-7206-2671}{\includegraphics[scale=0.06]{orcid.pdf}\hspace{1mm}Jeroen Bertels}\footnotemark[1]\\
	Processing Speech and Images\\
	Department of Electrical Engineering\\
	KU Leuven, Belgium\\
	\texttt{jeroen.bertels@kuleuven.be}\\
	\And
	David Robben\\
	icometrix\\
	Kolonel Begaultlaan 1b/12\\
	Leuven, Belgium\\
	\texttt{david.robben@kuleuven.be}\\
	\And
	Dirk Vandermeulen\\
	Processing Speech and Images\\
	Department of Electrical Engineering\\
	KU Leuven, Belgium\\
	\texttt{dirk.vandermeulen@kuleuven.be}\\
    \And
	Frederik Maes\\
	Processing Speech and Images\\
	Department of Electrical Engineering\\
	KU Leuven, Belgium\\
	\texttt{frederik.maes@kuleuven.be}
}
\begin{document}
\date{}  % comment for current date; leave blank for no date
\maketitle
\begin{abstract}
Albeit the Dice loss is one of the dominant loss functions in medical image segmentation, most research omits a closer look at its derivative, i.e. the real motor of the optimization when using gradient descent. In this paper, we highlight the peculiar action of the Dice loss in the presence of missing or empty labels. First, we formulate a theoretical basis that gives a general description of the Dice loss and its derivative. It turns out that the choice of the reduction dimensions $\Phi$ and the smoothing term $\epsilon$ is non-trivial and greatly influences its behavior. We find and propose heuristic combinations of $\Phi$ and $\epsilon$ that work in a segmentation setting with either missing or empty labels. Second, we empirically validate these findings in a binary and multiclass segmentation setting using two publicly available datasets. We confirm that the choice of $\Phi$ and $\epsilon$ is indeed pivotal. With $\Phi$ chosen such that the reductions happen over a single batch (and class) element and with a negligible $\epsilon$, the Dice loss deals with missing labels naturally and performs similarly compared to recent adaptations specific for missing labels. With $\Phi$ chosen such that the reductions happen over multiple batch elements or with a heuristic value for $\epsilon$, the Dice loss handles empty labels correctly. We believe that this work highlights some essential perspectives and hope that it encourages researchers to better describe their exact implementation of the Dice loss in future work.
\end{abstract}
\section{Introduction}\label{sec:introduction}
The \textit{Dice loss} was introduced in~\cite{Drozdzal2016} and~\cite{Milletari2016a} as a loss function for binary image segmentation taking care of the class imbalance between foreground and background often present in medical applications. The \textit{generalized Dice loss}~\cite{Sudre2017} extended this idea to multiclass segmentation tasks, thereby taking into account the class imbalance that is present across different classes. In parallel, the Jaccard loss was introduced in the wider computer vision field for the same purpose~\cite{Tarlow2012,Nowozin2014}. More recently, it has been shown that one can use either Dice or Jaccard loss during training to effectively optimize both metrics at test time~\cite{Eelbode2020}.\\
The use of the Dice loss in popular and state-of-the-art methods such as No New-Net~\cite{Isensee2018} has only fueled its dominant usage across the entire field of medical image segmentation. Despite its fast and wide adoption, research that explores the underlying mechanisms is remarkably limited and mostly focuses on the loss value itself building further on the concept of \textit{risk minimization}~\cite{Goodfellow2016}. Regarding model calibration and inherent uncertainty, for example, some intuitions behind the typical hard and poorly calibrated predictions were exposed in~\cite{Bertels2021}, thereby focusing on the potential volume bias as a result of using the Dice loss. Regarding semi-supervised learning, adaptations to the original formulations were proposed to deal with ``missing'' labels \cite{Fidon2021a,Shi2021}, i.e. a label that is missing in the ground truth even though it is present in the image.\\
In this work, we further contribute to a deeper understanding of the specific implementation of the Dice loss, especially in the context of missing and empty labels. In contrast to missing labels, ``empty'' labels are labels that are not present in the image (and hence also not in the ground truth). We will first take a closer look at the derivative, i.e. the real motor of the underlying optimization when using gradient descent, in Sect.~\ref{sec:theory}. Although~\cite{Milletari2016a} and~\cite{Sudre2017} report the derivative, it is not being discussed in detail, nor is any reasoning behind the choice of the reduction dimensions $\Phi$ given (Sect. 2.1). When the smoothing term $\epsilon$ is mentioned, no details are given and its effect is underestimated by merely linking it with numerical stability~\cite{Sudre2017} and convergence issues~\cite{Isensee2018}. In fact, we find that both $\Phi$ and $\epsilon$ are intertwined, and that their choice is non-trivial and pivotal in the presence of missing or empty labels. To confirm and validate these findings, we set up two empirical settings with missing or empty labels in Sect.~\ref{sec:setup} and~\ref{sec:results}. Indeed, we can make or break the segmentation task depending on the exact implementation of the Dice loss.
\section{Bells and whistles of the Dice loss: $\Phi$ and $\epsilon$}\label{sec:theory}
\input{reductions_figure}
In a CNN-based setting, the weights $\theta\in\Theta$ are often updated using gradient descent. For this purpose, the loss function $\ell$ computes a real valued cost $\ell(Y,\tilde{Y})$ based on the comparison between the ground truth $Y$ and its prediction $\tilde{Y}$ in each iteration. $Y$ and $\tilde{Y}$ contain the values $y_{b,c,i}$ and $\tilde{y}_{b,c,i}$, respectively, pointing to the value for a semantic class $c\in\mathcal{C}=[\text{C}]$ at an index $i\in\mathcal{I}=[\text{I}]$ (e.g. a voxel) of a batch element $b\in\mathcal{B}=[\text{B}]$ (Figure~\ref{fig:reductions}). The exact update of each $\theta$ depends on $d\ell(Y,\tilde{Y})/d\theta$, which can be computed via the generalized chain rule. With $\omega=(b,c,i)\in\Omega=\mathcal{B} \times \mathcal{C} \times \mathcal{I}$, we can write:
\begin{equation}
    \frac{d\ell(Y,\tilde{Y})}{d\theta}= 
    \sum_{b\in\mathcal{B}}\sum_{c\in\mathcal{C}}\sum_{i\in\mathcal{I}}\frac{\partial\ell(Y,\tilde{Y})}{\partial\tilde{y}_{b,c,i}}
    \frac{\partial\tilde{y}_{b,c,i}}{\partial\theta}=
    \sum_{\omega\in\Omega}\frac{\partial\ell(Y,\tilde{Y})}{\partial\tilde{y}_\omega}
    \frac{\partial\tilde{y}_\omega}{\partial\theta}.
    \label{eq:sum_of_partials}
\end{equation}
The Dice similarity coefficient (DSC) over a subset $\phi\subset\Omega$ is defined as:
\begin{equation}
    \text{DSC}(Y_\phi,\tilde{Y}_\phi)=\frac{2|Y_\phi\cap\tilde{Y}_\phi|}{|Y_\phi|+|\tilde{Y}_\phi|}.
\end{equation}
This formulation of $\text{DSC}(Y_\phi,\tilde{Y}_\phi)$ requires $Y$ and $\tilde{Y}$ to contain values in $\{0, 1\}$. In order to be differentiable and handle values in $[0, 1]$, relaxations such as the \textit{soft} DSC (sDSC) are used \cite{Drozdzal2016,Milletari2016a}. Furthermore, in order to allow both $Y$ and $\tilde{Y}$ to be empty, a smoothing term $\epsilon$ is added to the nominator and denominator such that $\text{DSC}(Y_\phi,\tilde{Y}_\phi)=1$ in case both $Y$ and $\tilde{Y}$ are empty. This results in the more general formulation of the Dice loss (DL) computed over a number of subsets $\Phi = \{\phi\}$:
\begin{equation}
    \text{DL}(Y,\tilde{Y})=
    1-\frac{1}{|\Phi|}\sum_{\phi\in\Phi}\text{sDSC}(Y_\phi,\tilde{Y}_\phi)=
    1-\frac{1}{|\Phi|}\sum_{\phi\in\Phi}\frac{2\sum_{\varphi\in\phi} y_\varphi\tilde{y}_\varphi+\epsilon}{\sum_{\varphi\in\phi} (y_\varphi+\tilde{y}_\varphi)+\epsilon}.
    \label{eq:soft_dice}
\end{equation}
Note that typically all $\phi$ are equal in size and define a partition over the domain $\Omega$, such that $\bigcup_{\phi\in\Phi}\phi=\Omega$ and $\bigcap_{\phi\in\Phi}\phi=0$. In $d\text{DL}(Y,\tilde{Y})/d\theta$ from Eq.~\ref{eq:sum_of_partials}, the derivative $\partial \text{DL}(Y,\tilde{Y})/\partial\tilde{y}_\omega$ acts as a scaling factor. In order to understand the underlying optimization mechanisms we can thus analyze $\partial \text{DL}(Y,\tilde{Y})/\partial\tilde{y}_\omega$. Given that all $\phi$ are disjoint, this can be written as:
\begin{equation}
    \frac{\partial \text{DL}(Y,\tilde{Y})}{\partial\tilde{y}_\omega}    =-\frac{1}{|\Phi|}\left(\frac{2y_\omega}{\sum_{\varphi\in\phi^\omega}(y_\varphi+\tilde{y}_\varphi)+\epsilon}-
\frac{2\sum_{\varphi\in\phi^\omega}y_\varphi\tilde{y}_\varphi+\epsilon}{\left(\sum_{\varphi\in\phi^\omega}(y_\varphi+\tilde{y}_\varphi)+\epsilon\right)^2}\right),
    \label{eq:gradient}
\end{equation}
with $\phi^\omega$ the subset that contains $\omega$. As such, it becomes clear that the specific action of DL depends on the exact configuration of the partition $\Phi$ of $\Omega$ and the choice of $\epsilon$. Next, we describe the most common choices of $\Phi$ and $\epsilon$ in practice. Then, we investigate their effects in the context of missing or empty labels. Finally, we present a simple heuristic to tune both.%, depending on the exact situation.
\subsection{Configuration of $\Phi$ and $\epsilon$ in practice}\label{sec:configurations}
In Figure~\ref{fig:reductions}, we depict four straightforward choices for $\Phi$. We define these as the \textit{image-wise}, \textit{class-wise}, \textit{batch-wise} or \textit{all-wise} DL implementation, respectively $\text{DL}_\mathbb{I}$, $\text{DL}_\mathbb{CI}$, $\text{DL}_\mathbb{BI}$ and $\text{DL}_\mathbb{BCI}$, thus referring to the dimensions over which a complete reduction (i.e. the summations $\sum_{\varphi\in\phi}$ in Eq.~\ref{eq:soft_dice} and Eq.~\ref{eq:gradient}) is performed. We see that in all cases, a complete reduction is performed over the set of image indices $\mathcal{I}$, which is in line with all relevant literature that we consulted. Furthermore, while in most implementations $\text{B}>1$, only in~\cite{Kodym2019} the exact usage of the batch dimension is described. In fact, they experimented with both $\text{DL}_\mathbb{I}$ and $\text{DL}_\mathbb{BI}$, and found the latter to be superior for head and neck organs at risk segmentation in radiotherapy. Based on the context, we assume that most other contributions \cite{Drozdzal2016,Eelbode2020,Isensee2018,Jadon2020,Milletari2016a,Yeung2022} used $\text{DL}_\mathbb{I}$, although we cannot rule out the use of $\text{DL}_\mathbb{BI}$. Similarly, we assume that in~\cite{Sudre2017} $\text{DL}_\mathbb{CI}$ was used (with additionally weighting the contribution of each class inversely proportional to the object size), although we cannot rule out the use of $\text{DL}_\mathbb{BCI}$.\\
Note that in Eq.~\ref{eq:soft_dice} and Eq.~\ref{eq:gradient} we have assumed the choice for $\Phi$ and $\epsilon$ to be fixed. As such, the loss value or gradients only vary across different iterations due to a different sampling of $Y$ and $\tilde{Y}$. Relaxing this assumption allows us to view the \textit{leaf Dice loss} from~\cite{Fidon2021a} as a special case of choosing $\Phi$. Being developed in the context of missing labels, the partition $\Phi$ of $\Omega$ is altered each iteration by substituting each $\phi$ with $\emptyset$ if $\sum_\varphi^\phi y_\varphi=0$. Similarly, the \textit{marginal Dice loss} from~\cite{Shi2021} adapts $\Phi$ every iteration by treating the missing labels as background and summing the predicted probabilities of unlabeled classes to the background prediction before calculating the loss.\\  
Based on our own experience, $\epsilon$ is generally chosen to be small (e.g. $10^{-7}$). However, most research does not include $\epsilon$ in their loss formulation, nor do they mention its exact value. We do find brief mentions related to convergence issues~\cite{Isensee2018} (without further information) or numerical stability in the case of empty labels~\cite{Jadon2020,Sudre2017} (to avoid division by zero in Eq.~\ref{eq:soft_dice} and Eq.~\ref{eq:gradient}).
\subsection{Effect of $\Phi$ and $\epsilon$ on missing or empty labels}\label{sec:effects}
When inspecting the derivative given in Eq.~\ref{eq:gradient}, we notice that in a way $\partial \text{DL}/\partial\tilde{y}_\omega$ does not depend on $\tilde{y}_\omega$ itself. Instead, the contributions of $\tilde{y}_\varphi$ are aggregated over the reduction dimensions, resulting in a global effect of prediction $\tilde{Y}_\phi$. Consequently, the derivative in a subset $\phi$ takes only two distinct values corresponding to $y_\omega=0$ or $y_\omega=1$. This is in contrast to the derivative shown in~\cite{Milletari2016a} who used a $L^2$ norm-based relaxation, which causes the gradients to be different for every $\omega$ if $\tilde{y}_\omega$ is different. If we work further with the $L^1$ norm-based relaxation (following the vast majority of implementations) and assuming that $\sum_{\varphi\in\phi^\omega}\tilde{y}_\varphi\gg\epsilon$, we see that $\partial \text{DL}/\partial\tilde{y}_\omega$ will be negligible for missing or empty ground truth labels. Exploiting this property, we can either avoid having to implement specific losses for missing labels, or we can learn to predict empty maps with a good configuration of $\Phi$. Regarding the former, we simply need to make sure $\sum_{\varphi\in\phi^\omega}y_{\varphi}=0$ for each map that contains missing labels which can be achieved by using the image-wise implementation $\text{DL}_{\mathbb{I}}$. Regarding the latter, non-zero gradients are required for empty maps. Hence, we want to choose $\phi$ large enough to avoid $\sum_{\varphi\in\phi^\omega}y_{\varphi}=0$ for which a batch-wise implementation $\text{DL}_{\mathbb{BI}}$ is suitable.
\subsection{A simple heuristic for tuning $\epsilon$ to learn from empty maps}\label{sec:heuristic}
We hypothesized that we can learn to predict empty maps by using the batch-wise implementation $\text{DL}_\mathbb{BI}$. However, due to memory constraints and trade-off with receptive field, it is often not possible to go for large batch sizes. In the limits when $\text{B}=1$ we find that $\text{DL}_\mathbb{I}=\text{DL}_\mathbb{BI}$, and thus the gradients of empty maps will be negligible. Hence, we want to mimic the behavior of $\text{DL}_{\mathbb{BI}}$ with $\text{B}\gg 1$, but using $\text{DL}_{\mathbb{I}}$. This can be achieved by tuning $\epsilon$ to increase the derivative for empty labels $y_\omega=0$. A very simple strategy would be to let $\partial \text{DL}(Y,\tilde{Y})/\partial\tilde{y}_\omega$ for $y_\omega=0$ be equal in case of (i) $\text{DL}_\mathbb{BI}$ with infinite batch size such that $\sum_{\varphi\in\phi^\omega}y_{\varphi} \neq 0$ and negligible $\epsilon$ and (ii) $\text{DL}_\mathbb{I}$ with non-negligible epsilon and $\sum_{\varphi\in\phi^\omega}y_\varphi=0$. If we set $\sum_{\varphi\in\phi^\omega}\tilde{y}_\varphi=\hat{v}$ we get: 
\begin{equation}
\frac{2\sum_{\varphi\in\phi^\omega}y_\varphi\tilde{y}_\varphi}{\left(\sum_{\varphi\in\phi^\omega}(y_\varphi+\tilde{y}_\varphi)\right)^2}=\frac{\epsilon}{\left(\sum_{\varphi\in\phi^\omega}\tilde{y}_\varphi+\epsilon\right)^2} \Rightarrow  \frac{2a\hat{v}}{(b\hat{v})^2}=\frac{\epsilon}{(\hat{v}+\epsilon)^2},
\label{eq:eps}
\end{equation}
with $a$ and $b$ variables to express the intersection and union as a function of $\hat{v}$. We can easily see that when we assume the overlap to be around 50~\%, thus $a\approx1/2$, and $\sum_{\varphi\in\phi^\omega}y_\varphi\approx\sum_{\varphi\in\phi^\omega} \tilde{y}_\varphi=\hat{v}$, thus $b\approx2$, we can find $\epsilon\approx\hat{v}$. It is further reasonable to assume that after some iterations $\hat{v}\approx\mathbb{E}\sum_{\varphi\in\phi^\omega} y_\varphi$, thus setting $\epsilon = \hat{v}$ will allow DL to learn empty maps.
\section{Experimental setup}\label{sec:setup}
To confirm empirically the observed effects of $\Phi$ and $\epsilon$ on missing or empty labels (Sect.~\ref{sec:effects}), and to test our simple heuristic choice of $\epsilon$ (Sect.~\ref{sec:heuristic}), we perform experiments using three implementations of DL on two different public datasets.

\textbf{Setups $\mathbb{I}$, $\mathbb{BI}$ and $\mathbb{I}_\epsilon$:} In $\mathbb{I}$ and $\mathbb{BI}$, respectively $\text{DL}_\mathbb{I}$ and $\text{DL}_\mathbb{BI}$ are used to calculate the Dice loss (Sect.~\ref{sec:configurations}). The difference between $\mathbb{I}$ and $\mathbb{I}_\epsilon$ is that we use a negligible value for epsilon $\epsilon=10^{-7}$ in $\mathbb{I}$ and use the heuristic from Sect.~\ref{sec:heuristic} to set $\epsilon=\mathbb{E}\sum_{\varphi\in\phi^\omega} y_\varphi$ in $\mathbb{I}_\epsilon$. From Sect.~\ref{sec:effects}, we expect $\mathbb{I}$ (any B) and $\mathbb{BI}$ ($\text{B}=1$) to successfully ignore missing labels during training, still segmenting these at test time. Vice versa, we expect $\mathbb{BI}$ ($\text{B}>1$) and $\mathbb{I}_\epsilon$ (any B) to successfully learn what maps should be empty and thus output empty maps at test time.

\textbf{BRATS:} For our purpose, we resort to the binary segmentation of whole brain tumors on pre-operative MRI in BRATS 2018~\cite{Menze2015,Bakas2017,Bakas2018a}. The BRATS 2018 training dataset consists of 75 subjects with a lower grade glioma (LGG) and 210 subjects with a glioblastoma (HGG). To construct a partially labeled dataset for the missing and empty label tasks, we substitute the ground truth segmentations of the LGGs with empty maps during training.  In light of missing labels, we would like the CNN to successfully segment LGGs at test time. In light of empty maps, we would like the CNN to output empty maps for LGGs at test time. Based on the ground truths of the entire dataset, in $\mathbb{I}_\epsilon$ we need to set $\epsilon=8,789$ or $\epsilon=12,412$ when we use the partially or fully labeled dataset for training, respectively.

\textbf{ACDC:} The ACDC dataset~\cite{Bernard2018} consists of cardiac MRI of 100 subjects. Labels for left ventricular (LV) cavity, LV myocardium and right ventricle (RV) are available in end-diastole (ED) and end-systole (ES). To create a structured partially labeled dataset, we remove the myocardium labels in ES. This is a realistic scenario since segmenting the myocardium only in ED is common in clinical practice. More specifically, ED and ES were sampled in the ratio 3/1 for $\mathbb{I}_\epsilon$, resulting in $\epsilon$ being equal to 13,741 and 19,893 on average for the myocardium class during partially or fully labeled training, respectively. For LV and RV, $\epsilon$ was 21,339 and 18,993, respectively. We ignored the background map when calculating DL. Since we hypothesize that $\text{DL}_\mathbb{I}$ is able to ignore missing labels, we compare $\mathbb{I}$ to the marginal Dice loss \cite{Shi2021} and the leaf Dice loss \cite{Fidon2021a}, two loss functions designed in particular to deal with missing labels. 

\textbf{Implementation details:} We start from the exact same preprocessing, CNN architecture and training parameters as in No New-Net~\cite{Isensee2018}.
The images of the BRATS dataset were first resampled to an isotropic voxel size of 2~x~2~x 2~mm$^3$, such that we could work with a smaller output segment size of 80~x~80~x 48~voxels as to be able to vary B in $\{1, 2, 4, 8\}$. Since we are working with a binary segmentation task we have $\text{C}=1$ and use a single sigmoid activation in the final layer.
For ACDC, the images were first resampled to 192~x~192~x~48 with a voxel size of 1.56~x~1.56 x~2.5~mm$^3$. The aforementioned CNN architecture was modified to use batch normalization and pReLU activations. To compensate the anisotropic voxel size, we used a combination of 3~x~3~x~3 and 3~x~3~x~1 convolutions and omitted the first max-pooling for the third dimension. These experiments were only performed for $\text{B}=2$. In this multiclass segmentation task, we use a softmax activation in the final layer to obtain four output maps.
\textbf{Statistical performance:} 
All experiments were performed under a five-fold cross-validation scheme, making sure each subject was only present in one of the five partitions. Significant differences were assessed with non-parametric bootstrapping, making no assumptions on the distribution of the results~\cite{Bakas2018a}. Results were considered statistically significant if the p-value was below 5~\%.
\section{Results}\label{sec:results}
Table~\ref{tab:I_IB_Ie} reports the mean DSC and mean volume difference ($\mathrm{\Delta}$V) between the fully labeled validation set and the predictions for tumor (BRATS) and myocardium (ACDC). For both the label that was always available (HGG or MYO\textsubscript{ED}) and the label that was not present in the partially labeled training dataset (LGG or MYO\textsubscript{ES}), we can make two observations. First, configurations $\mathbb{I}$ and $\mathbb{BI}$ ($\text{B}=1$) delivered a comparable segmentation performance (in terms of both DSC and $\mathrm{\Delta}$V) compared to using a fully labeled training dataset. Second, using configurations $\mathbb{BI}$ ($\text{B}>1$) and $\mathbb{I}_\epsilon$ the performance was consistently inferior. In this case, the CNN starts to learn when it needs to output empty maps. As a result, when calculating the DSC and $\mathrm{\Delta}$V with respect to a fully labeled validation dataset, we expect both metrics to remain similar for HGG and MYO\textsubscript{ES}. On the other hand, we expect a mean DSC of 0 and a $|\mathrm{\Delta}$V$|$ close to the mean volume of LGG or MYO\textsubscript{ES}. Note that this is not the case due to the incorrect classification of LGG or MYO\textsubscript{ES} as HGG or MYO\textsubscript{ED}, respectively. Figure~\ref{fig:roc_figures} shows the Receiver Operating Characteristic (ROC) curves when using a partially labeled training dataset with the goal to detect HGG or MYO\textsubscript{ED} based on a threshold on the predicted volume at test time. For both tasks, we achieved an Area Under the Curve (AUC) of around 0.9. Figure~\ref{fig:qualitative_examples} shows an example segmentation.\\
When comparing $\mathbb{I}$ with the marginal Dice loss~\cite{Shi2021} and the leaf Dice loss~\cite{Fidon2021a}, no significant differences between any method for myocardium (MYO\textsubscript{ED}~=~0.88, MYO\textsubscript{ES}~= 0.88), LV (LV\textsubscript{ED}~=~0.96, LV\textsubscript{ES}~=~0.92) and RV (RV\textsubscript{ED}~=~0.93, RV\textsubscript{ES}~=~0.86~-~0.87) were found in both ED and ES.
\input{results_table_true_empty}
\input{roc_figures}
\input{qualitative_examples}
\section{Discussion}
The experiments confirmed the analysis from Sect.~\ref{sec:effects} that $\text{DL}_{\mathbb{I}}$ (equal to $\text{DL}_{\mathbb{BI}}$ when $\text{B}=1$) ignores missing labels during training and that it can be used in the context of missing labels naively. On the other hand, we confirmed that $\text{DL}_{\mathbb{BI}}$ (with $\text{B}>1$) and  $\text{DL}_{\mathbb{I}}$ (with a heuristic choice of $\epsilon$) can effectively learn to predict empty labels, e.g. for classification purposes or to be used with small patch sizes.\\
When heuristically determining $\epsilon$ for configuring $\mathbb{I}_{\epsilon}$ (Eq.~\ref{eq:eps}), we only focused on the derivative for $y_\omega=0$. Of course, by adapting $\epsilon$, the derivative for $y_\omega=1$ will also change. Nonetheless, our experiments showed that $\mathbb{I}_\epsilon$ can achieve the expected behavior, indicating that the effect on the derivative for $y_\omega=1$ is only minor compared to $y_\omega=0$. We wish to derive a more exact formulation of the optimal value of $\epsilon$ in future work. We expect this optimal $\epsilon$ to depend on the distribution between the classes, object size and other labels that might be present. Furthermore, it would be interesting to study the transition between the near-perfect prediction for the missing class ($\text{DL}_\mathbb{I}$ with small $\epsilon$) and the prediction of empty labels for the missing class ($\text{DL}_\mathbb{I}$ with large $\epsilon$).\\
All the code necessary for exact replication of the results including preprocessing, training scripts, statistical analysis, etc. was released to encourage further analysis on this topic (\url{https://github.com/JeroenBertels/dicegrad}). 
\section{Conclusion}
We showed that the choice of the reduction dimensions $\Phi$ and the smoothing term $\epsilon$ for the Dice loss is non-trivial and greatly influences its behavior in the context of missing or empty labels. We believe that this work highlights some essential perspectives and hope that it encourages researchers to better describe their exact implementation of the Dice loss in the future.\\
\section*{Acknowledgements}
This research received funding from the Flemish Government under the “Onderzoeksprogramma Artificiële intelligentie (AI) Vlaanderen" programme and is also partially funded by KU Leuven Internal Funds C24/18/047 (F. Maes).
\bibliographystyle{plain}
\bibliography{dicelossmissinglabels,dicegrad}
\end{document}

%% file: reductions_figure.tex
\begin{figure}[b]
\setlength\tabcolsep{0pt}
\centering
\includegraphics[scale=0.8]{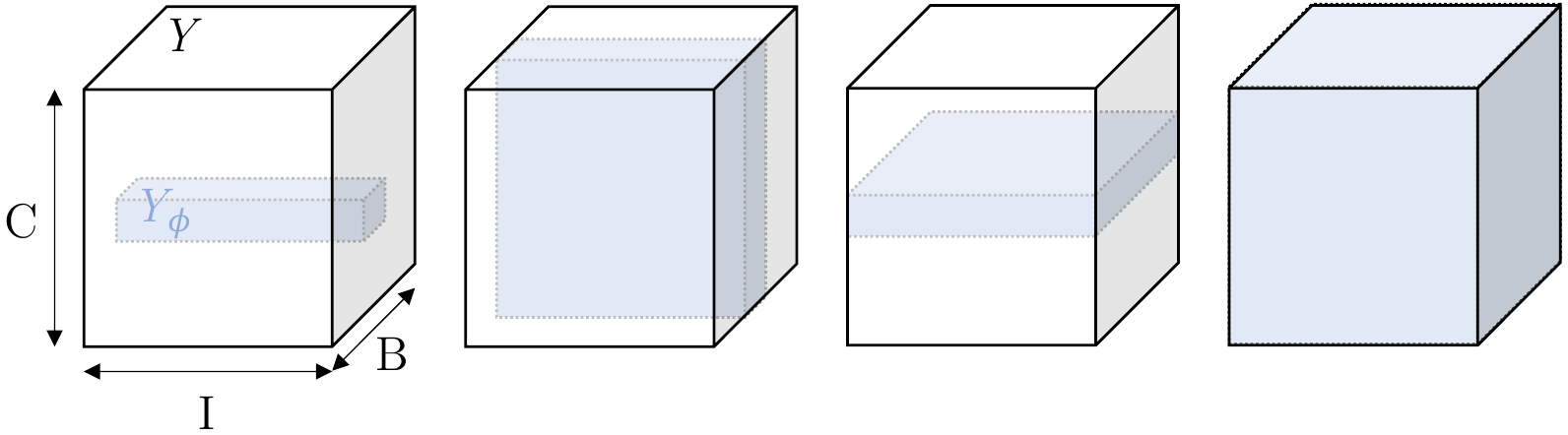}
\caption[Reductions.]{Schematic representation of $Y$, having a batch, class and image dimension, respectively with $|\mathcal{B}|=\text{B}$, $|\mathcal{C}|=\text{C}$ and $|\mathcal{I}|=\text{I}$ (similarly for $\tilde{Y}$). The choice of $\Phi$, i.e. a family of subsets $\phi$ over $\Omega$ defines the extent of the reductions in $\text{sDSC}(Y_\phi, \tilde{Y}_\phi)$. From left to right, we see how the choice of $\Phi$, and thus an example subset $\phi$ in blue, is different between the image-wise ($\text{DL}_\mathbb{I}$), class-wise ($\text{DL}_\mathbb{CI}$), batch-wise ($\text{DL}_\mathbb{BI}$) and all-wise ($\text{DL}_\mathbb{BCI}$) implementation of DL.}
\label{fig:reductions}
\end{figure}

%% file: results_table_true_empty.tex
\begin{table}[t]
\setlength\tabcolsep{3pt}
\centering
\caption[Results.]{Mean DSC and mean $\mathrm{\Delta}$V. HGG and MYO\textsubscript{ED} are always present during training while LGG and MYO\textsubscript{ES} are replaced by empty maps under partial labeling. Configurations that we expect to learn to predict empty maps are highlighted (since we used a fully labeled validation set, we expect lower DSC and $\mathrm{\Delta}$V). Comparing partial with full labeling, inferior ($\text{p}<0.05$) results are indicated in italic.}
\label{tab:I_IB_Ie}
\begin{tabular}{llc|ccc|ccc|ccc|ccc}
\toprule
&&& \multicolumn{6}{c|}{DSC} & \multicolumn{6}{c}{$\mathrm{\Delta}$V [ml]}\\
% \cline{4-9} \cline{10-15}
&&& \multicolumn{3}{c|}{HGG/MYO\textsubscript{ED}} & \multicolumn{3}{c|}{LGG/MYO\textsubscript{ES}}& \multicolumn{3}{c|}{HGG/MYO\textsubscript{ED}} & \multicolumn{3}{c}{LGG/MYO\textsubscript{ES}}\\
% \cline{4-15}
& Labeling & B & $\mathbb{I}$ & $\mathbb{BI}$ & $\mathbb{I}_\epsilon$ & $\mathbb{I}$ & $\mathbb{BI}$ & $\mathbb{I}_\epsilon$ &
            $\mathbb{I}$ & $\mathbb{BI}$ & $\mathbb{I}_\epsilon$ & $\mathbb{I}$ & $\mathbb{BI}$ & $\mathbb{I}_\epsilon$\\
\midrule
\multirow{8}{*}{BRATS} &\multirow{4}{*}{Full} & 1 &0.89 &     0.89 &         0.89 &     0.89 &     0.89 &         0.89 &       -4 &       -4 &           -7 &       -8 &       -7 &           -9  \\
  &          & 2 & 0.89 &     0.89 &         0.89 &     0.89 &     0.88 &         0.88 &       -5 &       -5 &           -7 &       -9 &      -10 &          -12  \\
   &         & 4 & 0.89 &     0.89 &         0.89 &     0.88 &     0.90 &         0.89 &       -6 &       -5 &           -7 &      -11 &       -7 &          -11  \\
    &        & 8 & 0.89 &     0.89 &         0.89 &     0.89 &     0.88 &         0.89 &      -6 &       -4 &           -6 &      -12 &      -10 &           -9  \\
\cline{2-15}
&\multirow{4}{*}{Partial} & 1 & 0.89 &     0.89 &         \textit{0.83} &     \textit{0.88} &     0.88 &         \cellcolor{gray!20}\textit{0.23} &      \textit{-5} &       \textit{-5} &          \textit{-12} &      \textit{-11} &      \textit{-12} &         \cellcolor{gray!20} \textit{-88}  \\
&            & 2 & 0.89 &     \textit{0.83} &         \textit{0.82} &     0.88 &     \cellcolor{gray!20}\textit{0.24} &         \cellcolor{gray!20}\textit{0.16} &      -6 &      \textit{-12} &          \textit{-13} &      \textit{-12} &      \cellcolor{gray!20}\textit{-89} &          \cellcolor{gray!20}\textit{-96}  \\
 &           & 4 & 0.89 &     \textit{0.82} &         \textit{0.83} &     0.88 &     \cellcolor{gray!20}\textit{0.20} &         \cellcolor{gray!20}\textit{0.20} &      \textit{-6} &      \textit{-12} &          \textit{-12} &      \textit{-15} &      \cellcolor{gray!20}\textit{-93} &          \cellcolor{gray!20}\textit{-94}  \\
  &          & 8 &  0.89 &     \textit{0.82} &         \textit{0.83} &     \textit{0.88} &     \cellcolor{gray!20}\textit{0.20} &         \cellcolor{gray!20}\textit{0.23} &      -6 &      \textit{-12} &          \textit{-12} &      -14 &      \cellcolor{gray!20}\textit{-94} &          \cellcolor{gray!20}\textit{-90} \\
\midrule
\multirow{2}{*}{ACDC}& Full&2&0.88&0.88&0.87&0.89&0.89&0.89&-1&0&-2&-3&0&-3\\
\cline{2-15}
&Partial&2&\textit{0.88}&\textit{0.80}&\textit{0.80}&\textit{0.88}&\cellcolor{gray!20}\textit{0.08}&\cellcolor{gray!20}\textit{0.06}&0 & \textit{-11} &\textit{-14}& \textit{-5} & \cellcolor{gray!20}\textit{-129}&\cellcolor{gray!20}\textit{-131}\\

\bottomrule
\end{tabular}
\end{table}

%% file: roc_figures.tex
\begin{figure}[b]
\setlength\tabcolsep{0pt}
\centering
\begin{tabular}{ccc}
    \includegraphics[scale=0.45]{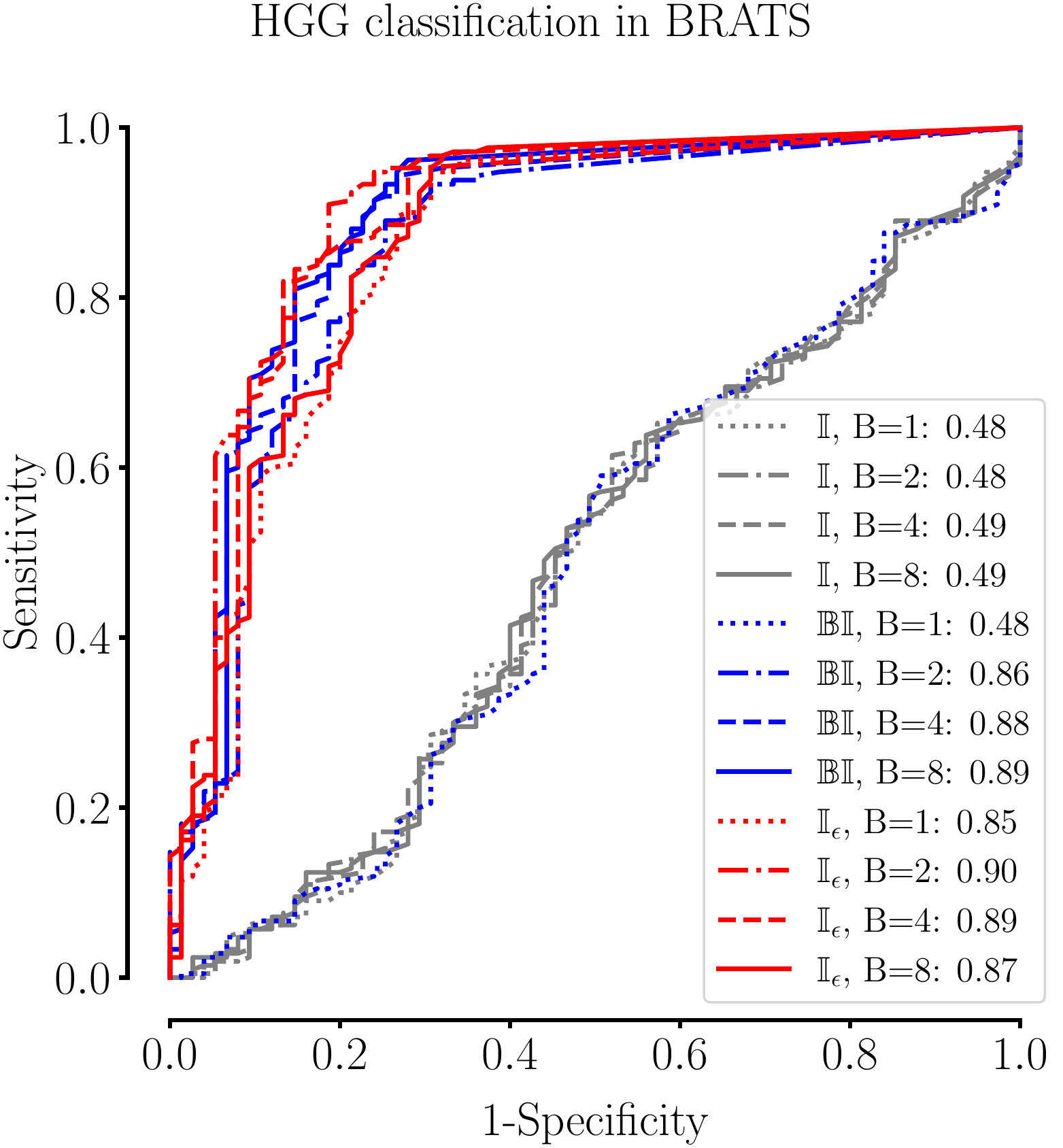}
    & ~~~~~~~
    &\includegraphics[scale=0.45]{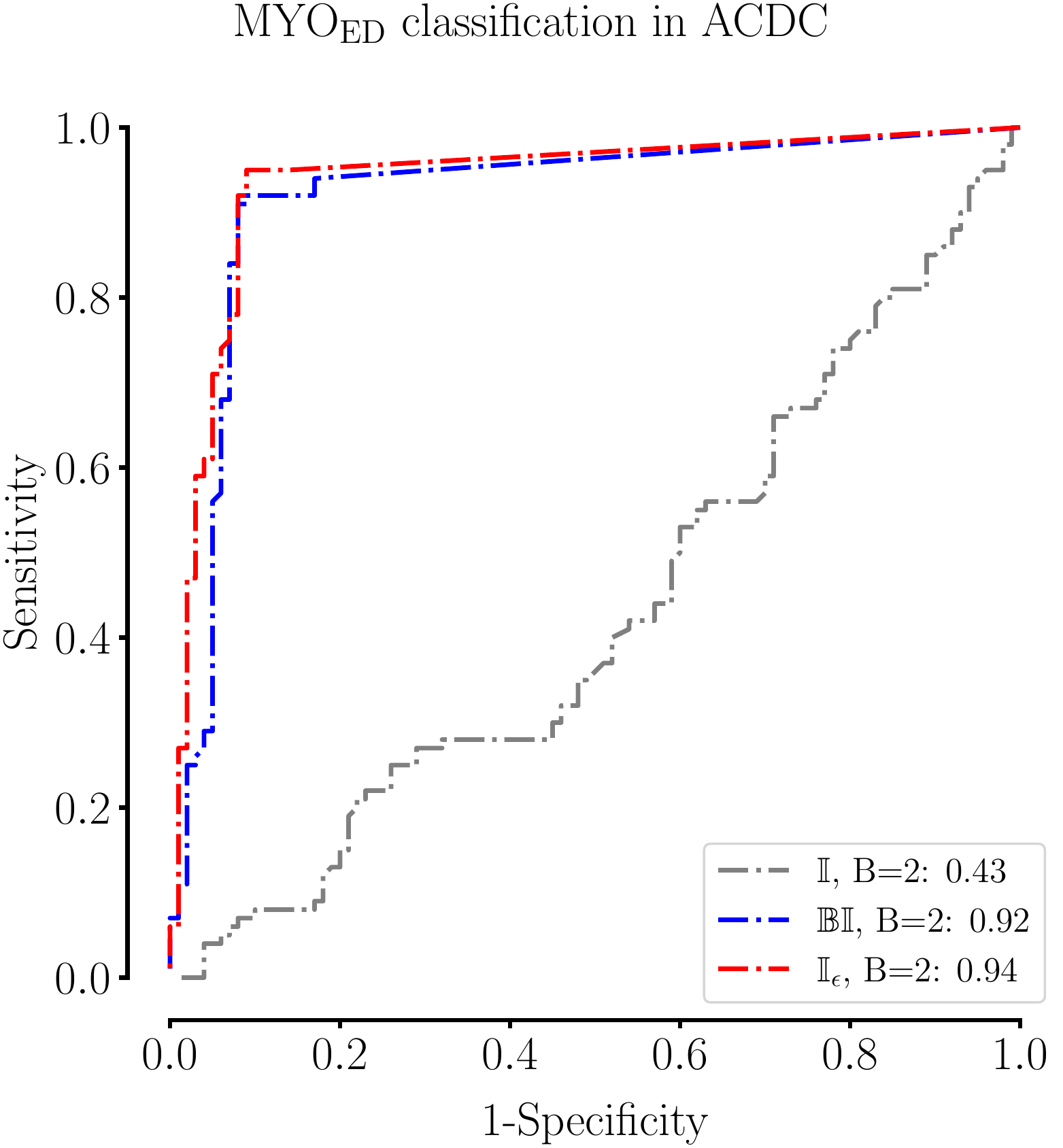}
\end{tabular}
\caption[ROC analysis.]{ROC analysis if we want to detect the label that was always present during training by using different thresholds on the predicted volume. In the legend we also report the AUC for each setting.}
\label{fig:roc_figures}
\end{figure}

%% file: qualitative_examples.tex
\begin{figure}[t]
\setlength\tabcolsep{0pt}
\centering

\newcommand\scale{0.55}
\begin{tabular}{cc|cccccccccc}
 &~& \multicolumn{4}{c}{HGG example}&~&~& \multicolumn{4}{c}{LGG example}\\
 &~& GT\textsubscript{train} & $\mathbb{I}$ & $\mathbb{BI}$ & $\mathbb{I}_\epsilon$ &~&~& GT\textsubscript{train} & $\mathbb{I}$ & $\mathbb{BI}$ & $\mathbb{I}_\epsilon$\\
\hline
\rotatebox[origin=c]{90}{$\text{B}=1$} &~
&\raisebox{-0.5\height}{\includegraphics[scale=\scale]{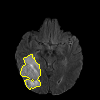}}
&\raisebox{-0.5\height}{\includegraphics[scale=\scale]{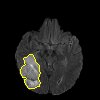}}
&\raisebox{-0.5\height}{\includegraphics[scale=\scale]{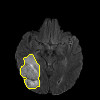}}
&\raisebox{-0.5\height}{\includegraphics[scale=\scale]{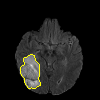}}&~&~
&\raisebox{-0.5\height}{\includegraphics[scale=\scale]{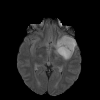}}
&\raisebox{-0.5\height}{\includegraphics[scale=\scale]{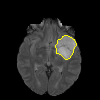}}
&\raisebox{-0.5\height}{\includegraphics[scale=\scale]{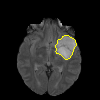}}
&\raisebox{-0.5\height}{\includegraphics[scale=\scale]{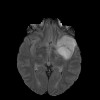}}\\
\rotatebox[origin=c]{90}{$\text{B}=2$} &~
&\raisebox{-0.5\height}{\includegraphics[scale=\scale]{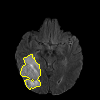}}
&\raisebox{-0.5\height}{\includegraphics[scale=\scale]{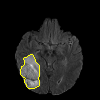}}
&\raisebox{-0.5\height}{\includegraphics[scale=\scale]{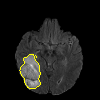}}
&\raisebox{-0.5\height}{\includegraphics[scale=\scale]{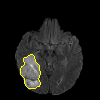}}&~&~
&\raisebox{-0.5\height}{\includegraphics[scale=\scale]{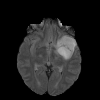}}
&\raisebox{-0.5\height}{\includegraphics[scale=\scale]{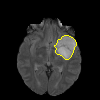}}
&\raisebox{-0.5\height}{\includegraphics[scale=\scale]{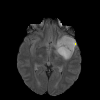}}
&\raisebox{-0.5\height}{\includegraphics[scale=\scale]{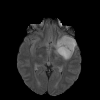}}\\
\rotatebox[origin=c]{90}{$\text{B}=4$} &~
&\raisebox{-0.5\height}{\includegraphics[scale=\scale]{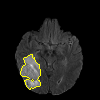}}
&\raisebox{-0.5\height}{\includegraphics[scale=\scale]{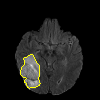}}
&\raisebox{-0.5\height}{\includegraphics[scale=\scale]{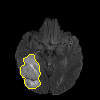}}
&\raisebox{-0.5\height}{\includegraphics[scale=\scale]{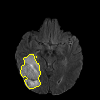}}&~&~
&\raisebox{-0.5\height}{\includegraphics[scale=\scale]{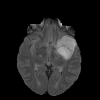}}
&\raisebox{-0.5\height}{\includegraphics[scale=\scale]{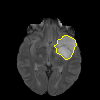}}
&\raisebox{-0.5\height}{\includegraphics[scale=\scale]{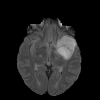}}
&\raisebox{-0.5\height}{\includegraphics[scale=\scale]{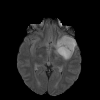}}\\
\rotatebox[origin=c]{90}{$\text{B}=8$} &~
&\raisebox{-0.5\height}{\includegraphics[scale=\scale]{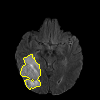}}
&\raisebox{-0.5\height}{\includegraphics[scale=\scale]{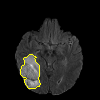}}
&\raisebox{-0.5\height}{\includegraphics[scale=\scale]{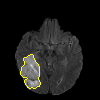}}
&\raisebox{-0.5\height}{\includegraphics[scale=\scale]{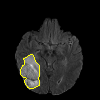}}&~&~
&\raisebox{-0.5\height}{\includegraphics[scale=\scale]{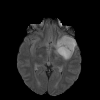}}
&\raisebox{-0.5\height}{\includegraphics[scale=\scale]{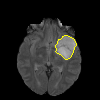}}
&\raisebox{-0.5\height}{\includegraphics[scale=\scale]{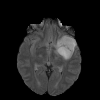}}
&\raisebox{-0.5\height}{\includegraphics[scale=\scale]{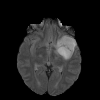}}\\
\multicolumn{9}{c}{\quad}
\end{tabular}

\renewcommand\scale{0.265}
\begin{tabular}{cc|cccccccccc}
&~ & \multicolumn{4}{c}{ED example}&~&~& \multicolumn{4}{c}{ES example}\\
&~ & GT\textsubscript{train} & $\mathbb{I}$ & $\mathbb{BI}$ & $\mathbb{I}_\epsilon$ &~&~& GT\textsubscript{train} & $\mathbb{I}$ & $\mathbb{BI}$ & $\mathbb{I}_\epsilon$\\
\hline
\rotatebox[origin=c]{90}{$\text{B}=2$} &~
&\raisebox{-0.5\height}{\includegraphics[scale=\scale]{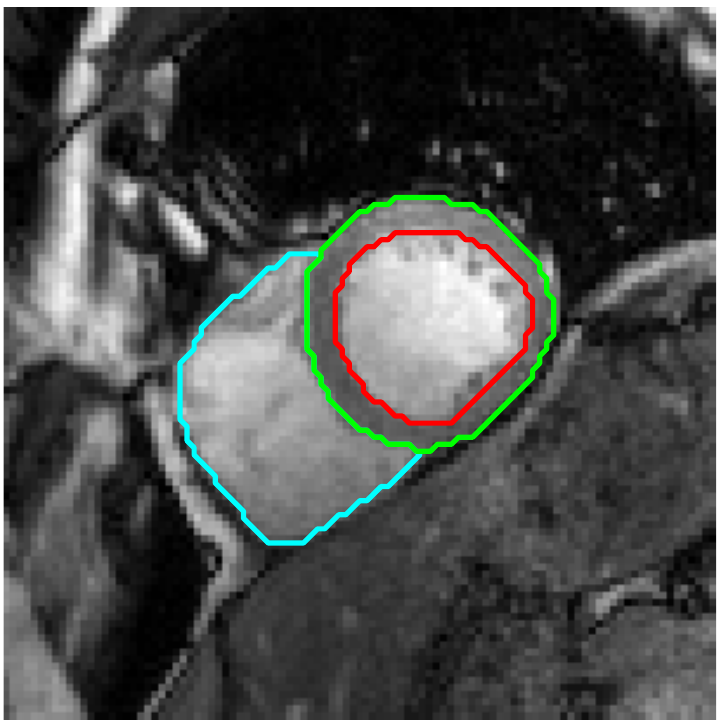}}
&\raisebox{-0.5\height}{\includegraphics[scale=\scale]{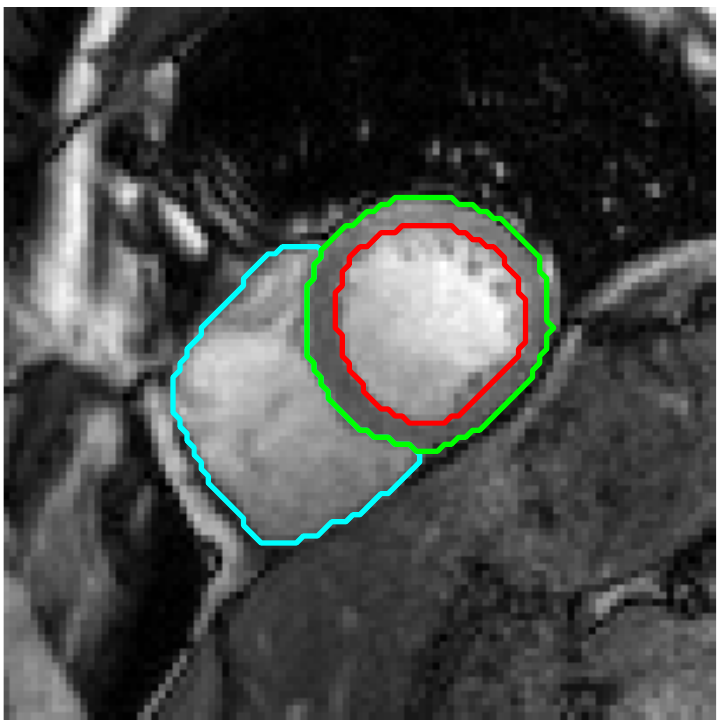}}
&\raisebox{-0.5\height}{\includegraphics[scale=\scale]{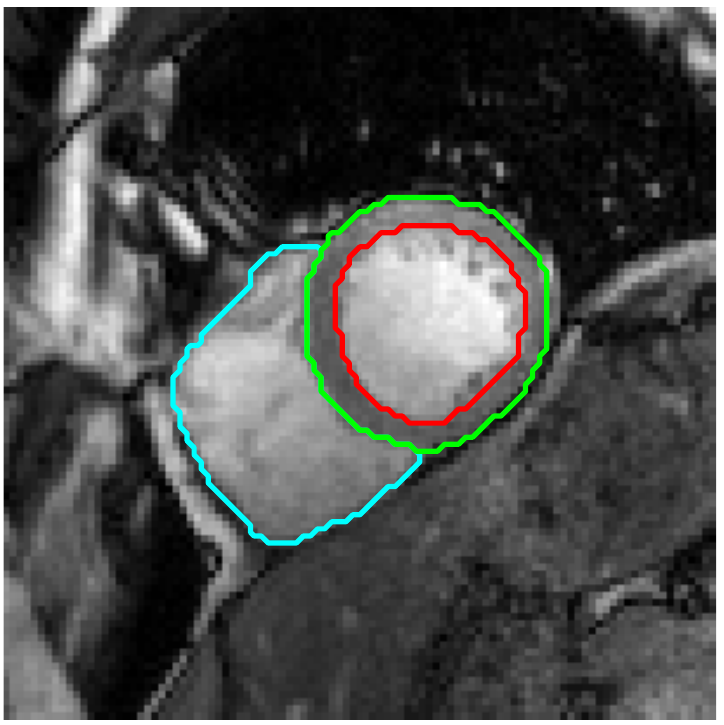}}
&\raisebox{-0.5\height}{\includegraphics[scale=\scale]{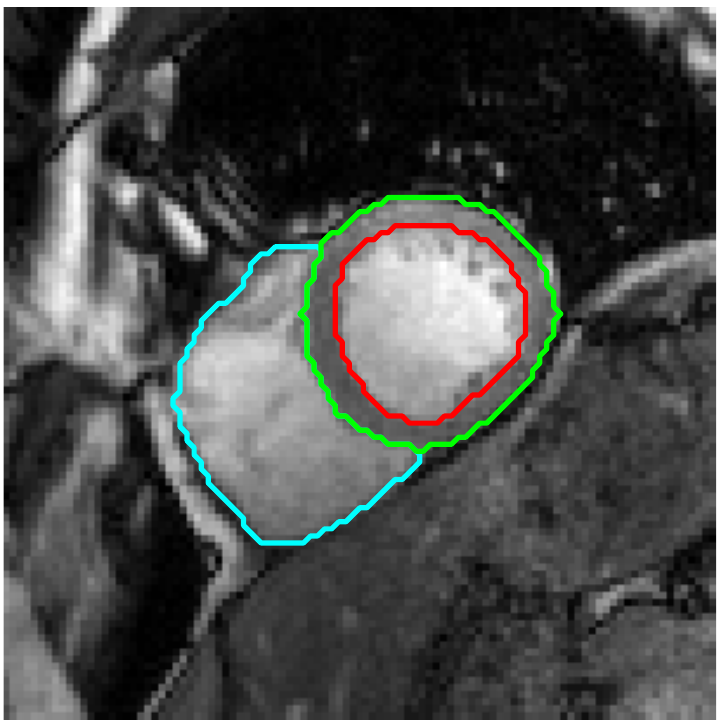}}&~&~
&\raisebox{-0.5\height}{\includegraphics[scale=\scale]{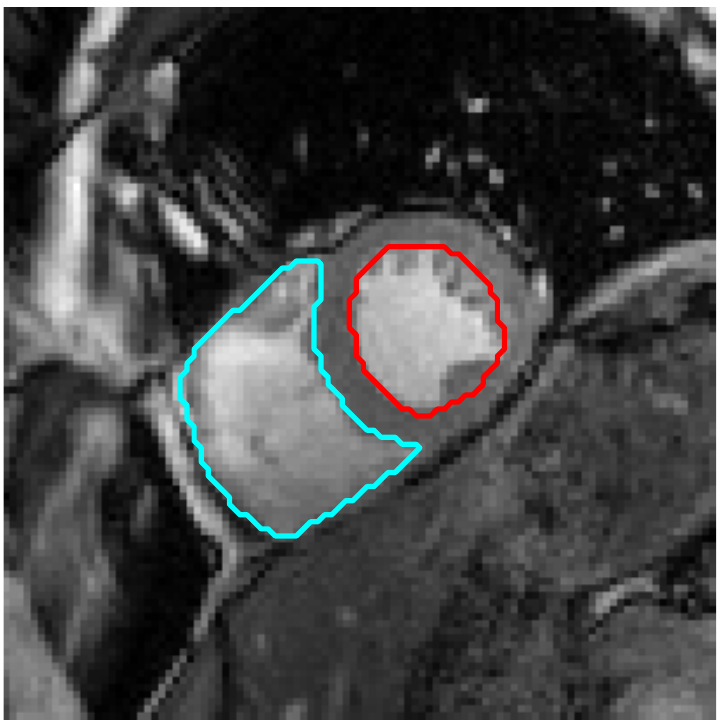}}
&\raisebox{-0.5\height}{\includegraphics[scale=\scale]{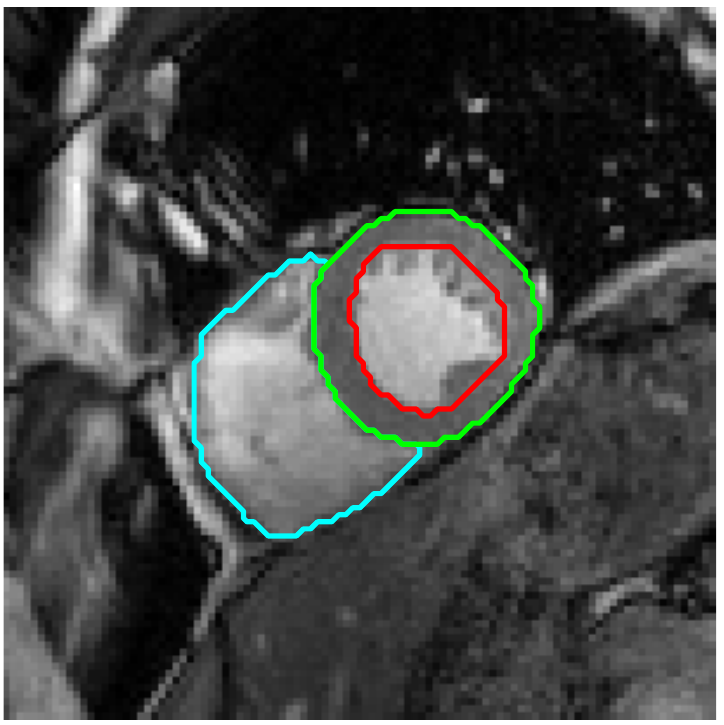}}
&\raisebox{-0.5\height}{\includegraphics[scale=\scale]{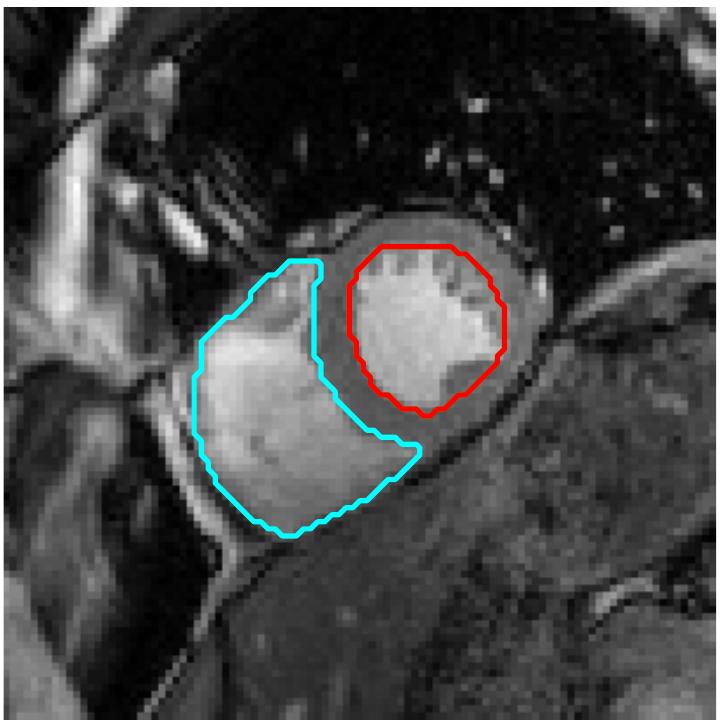}}
&\raisebox{-0.5\height}{\includegraphics[scale=\scale]{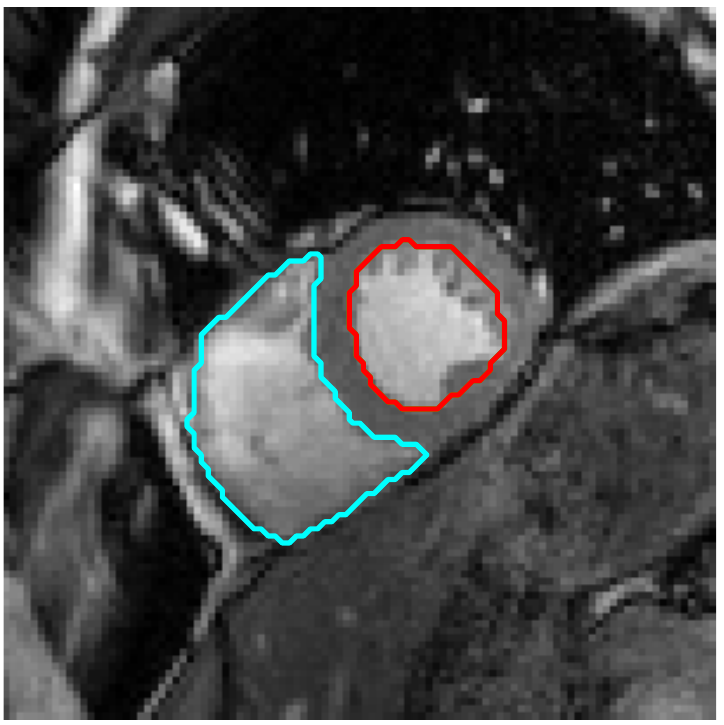}}
\end{tabular}

\caption[Qualitative examples.]{Segmentation examples for BRATS (top) and ACDC (bottom). The ground truths for LGG and MYO\textsubscript{ES} were replaced with empty maps during training (GT\textsubscript{train}).}
\label{fig:qualitative_examples}
\end{figure}